\documentclass[conference]{IEEEtran}
\IEEEoverridecommandlockouts

\usepackage{cite}
\usepackage{amsmath,amssymb,amsfonts}
\usepackage{algorithmic}
\usepackage{graphicx}
\usepackage{textcomp}
\usepackage{tabularx}
\usepackage{xcolor}
\usepackage{multirow}
\usepackage{adjustbox}
\usepackage{tabularx}
\usepackage{array}
\usepackage{colortbl}
\usepackage{booktabs}
\usepackage{stfloats}
\usepackage{lineno,hyperref}
\hypersetup{hidelinks,
	colorlinks=true,
	allcolors=black,
	pdfstartview=Fit,
	breaklinks=true}
\def\BibTeX{{\rm B\kern-.05em{\sc i\kern-.025em b}\kern-.08em
    T\kern-.1667em\lower.7ex\hbox{E}\kern-.125emX}}
\begin{document}

\title{Enhancing Few-Shot Out-of-Distribution Detection with Gradient Aligned Context Optimization\\

}

\author{\IEEEauthorblockN{1\textsuperscript{st} Baoshun Tong}
\IEEEauthorblockA{\textit{school of computer science and engineering} \\
\textit{sun yat-sen university}\\
Guangzhou, China \\
tongbsh@mail2.sysu.edu.cn}
\and
\IEEEauthorblockN{2\textsuperscript{nd} Kaiyu Song}
\IEEEauthorblockA{\textit{school of artificial intelligence} \\
\textit{sun yat-sen university}\\
Guangzhou, China \\
songky7@mail2.sysu.edu.cn}
\and
\IEEEauthorblockN{3\textsuperscript{rd} Hanjiang Lai$^{\ast}$}
\IEEEauthorblockA{\textit{school of computer science and engineering} \\
\textit{sun yat-sen university}\\
Guangzhou, China \\
laihanj3@mail.sysu.edu.cn}

}

\maketitle

\begin{abstract}
Few-shot out-of-distribution (OOD) detection aims to detect OOD images from unseen classes with only a few labeled in-distribution (ID) images. 
To detect OOD images and classify ID samples, prior methods have been proposed by regarding the background regions of ID samples as the OOD knowledge and performing OOD regularization and ID classification optimization.
However, the gradient conflict still exists between ID classification optimization and OOD regularization caused by biased recognition. 
To address this issue, we present \textbf{G}radient Aligned Context Optimization (GaCoOp) to mitigate this gradient conflict.
Specifically, we decompose the optimization gradient to identify the scenario when the conflict occurs.
Then we alleviate the conflict in inner ID samples and optimize the prompts via leveraging gradient projection. 
Extensive experiments over the large-scale ImageNet OOD detection benchmark demonstrate that our GaCoOp can effectively mitigate the conflict and achieve great performance. Code will be available at \href{https://github.com/BaoshunWq/ood_GaCoOp}{https://github.com/BaoshunWq/ood-GaCoOp}.
\end{abstract}


\section{Introduction}

Currently, deep learning has made tremendous progress, and most methods can achieve great performance when the training data and test data come from the same distribution. 
However, the test data does not always follow the same distribution as the training data~\cite{shu2023clipood}.
To address this issue, the researcher proposed a task named out-of-distribution (OOD) detection~\cite{hendrycks2016baseline}, which needs not only to classify distribution (ID) samples precisely but also to discern OOD samples accurately.

Owing to VLMs~\cite{radford2021learning} demonstrating promising performance in various downstream tasks, many researchers have been exploring its application in out-of-distribution (OOD) detection. Due to CLIP's~\cite{radford2021learning} excellent generalization ability, previous OOD detection works focus on detecting without utilizing any training samples, which is also known as zero-shot OOD detection~\cite{ming2022delving}. While some researchers~\cite{liang2018ODIN,wang2022vim,sun2022KNN,taonon-NPOS} pointed out that although the zero-shot methods do not require any training data, they may encounter a domain gap with ID downstream data.

Consequently, researchers~\cite{taonon-NPOS} have proposed fully supervised approaches that fine-tune the model using all available in-distribution training data. But~\cite{miyai2024locoop,bai2024id-like} believed that a fully supervised approach is computationally expensive and may compromise the excellent generalization ability inherent in CLIP itself.

To address the above challenges, a new task called few-shot out-of-distribution detection~\cite{miyai2024locoop} has been proposed. It performs out-of-distribution (OOD) detection using only a few in-distribution (ID) training data.
For example, LoCoOp~\cite{miyai2024locoop} is the representative work of few-shot out-of-distribution detection. According to the ability of CLIP to distinguish between background information and category information, LoCoOp regards background information as OOD information and prevents the model from producing high ID confidence scores for the OOD features via OOD regularization.

However, CLIP will lead the biased prediction~\cite{yang2023invariant}, where CLIP identifies irreverent background information as the main ID object (or vice versa). This causes a conflict between ID classification and OOD regularization optimization. Specifically, when CLIP can correctly distinguish category causal information and background information, there is almost no conflict here. 
But if CLIP incorrectly identifies category information as background information, the OOD regularization optimization will affect the ID classification and the performance of OOD detection~\cite{song2024fd-align}. Thus, CLIP's recognition bias for ID samples causes the gradient conflict~\cite{wang2024learnclip-bias}. 

To address this, we present a novel few-shot OOD detection prompt tuning method called \textbf{G}radient aligned \textbf{C}ontext \textbf{O}ptimization (GaCoOp). The principle of GaCoOp is to decouple the gradient of ID classification in each tuning step. Specifically, we measure the ID classification optimization direction $G_{i}$ using the gradient of cross-entropy classification loss between the ground-truth and the model's logits, which named as ID classification direction. Similarly, we compute the OOD regularization optimization direction $G_{o}$ using the gradient of OOD regularization loss in LoCoOp, dubbed OOD regularization direction. GaCoOp only updates the parameters in the optimization-aligned direction in each iteration with an acute angle to the OOD regularization direction.

We summarize our main contributions as follows: 
1) We propose GaCoOp, a simple yet effective OOD detection method to help mitigate the conflict between ID classification optimization and OOD regularization optimization due to CLIP's biased prediction. 
2) We find that the potential biases in CLIP can harm current OOD detection methods.
3) Experimental results show that our GaCoOp can improve performance on the large-scale ImageNet OOD benchmarks and increase the ID classification accuracy on ImageNet compared with the baseline.

\section{Related Work}
\label{sec:related}

\subsection{Vision-Language Models.} 

Generally, there are two types of architecture for vision-language models: single-stream models, exemplified by VisualBERT~\cite{li2019visualbert} and ViLT~\cite{kim2021vilt}, which feed concatenated text and visual features into a single transformer-based encoder; and dual-stream models like CLIP~\cite{radford2021learning} and FILIP~\cite{yaofilip}, which employ separate encoders for text and image, optimizing with contrastive objective to align semantically similar features across modalities. And due to CLIP has gained popularity thanks to its simplicity and robust performance, similar to prior works, we also adopt CLIP as the target pre-trained model.

\subsection{Prompt Learning}

The current methods to adapt VLMs~\cite{radford2021learning,li2022blip} with prompt can mainly be categorized into three types: language prompt learning~\cite{zhou2022CoOp}, visual prompt learning~\cite{jia2022visualprompt}, and multi-modal prompt learning~\cite{khattak2023maple}. In this paper, we mainly focus on language prompt learning. 
Currently, there are roughly three approaches to constructing language prompts. One involves manually crafting prompts~\cite{ming2022delving}, another entails generating personalized prompts that match the descriptions of target categories using large language models~\cite{pratt2023does}, and the final one utilizes a few training samples to train a learnable vector prompt~\cite{zhou2022CoOp}. This paper primarily focuses on the last category. CoOp~\cite{zhou2022CoOp} served as the representative work that significantly enhances CLIP’s performance on corresponding downstream tasks by learning the prompt from downstream data in the continual input embedding space. 
However, since these methods are proposed to address the issue of few-shot classification, they are not perfectly suited for the OOD detection task. So~\cite{miyai2024locoop} proposed LoCoOp, which can significantly improve performance on the OOD detection task with OOD regularization.

\subsection{Out-of-distribution detection}
There are a lot of works~\cite{hendrycks2016baseline,fort2021exploring,wang2022vim,du-vos,miyai2024locoop,bai2024id-like} that have been proposed to deal with the OOD detection task, where the test data has a different distribution from the train data due to label space shift. These methods can be divided into two categories~\cite{miyai2024locoop}. The first one is the common OOD detection approaches, such as probability-based method~\cite{liang2018ODIN}, feature-based method~\cite{wang2022vim}, and training-time regularization~\cite{du-vos}. The second one is the few-shot OOD detection methods that adapt foundational models such as CLIP to detect new classes. LoCoOp~\cite{miyai2024locoop} proposed to remove unnecessary information from the text embeddings of ID classes and prevents the model from producing high ID confidence scores for the OOD features with OOD regularization. ID-like~\cite{bai2024id-like} proposed a novel OOD detection framework that discovers ID-like outliers using CLIP from the vicinity space of the ID samples, thus helping to identify these most challenging OOD samples.

\begin{figure}[!ht]
        
        \includegraphics[width=\linewidth]{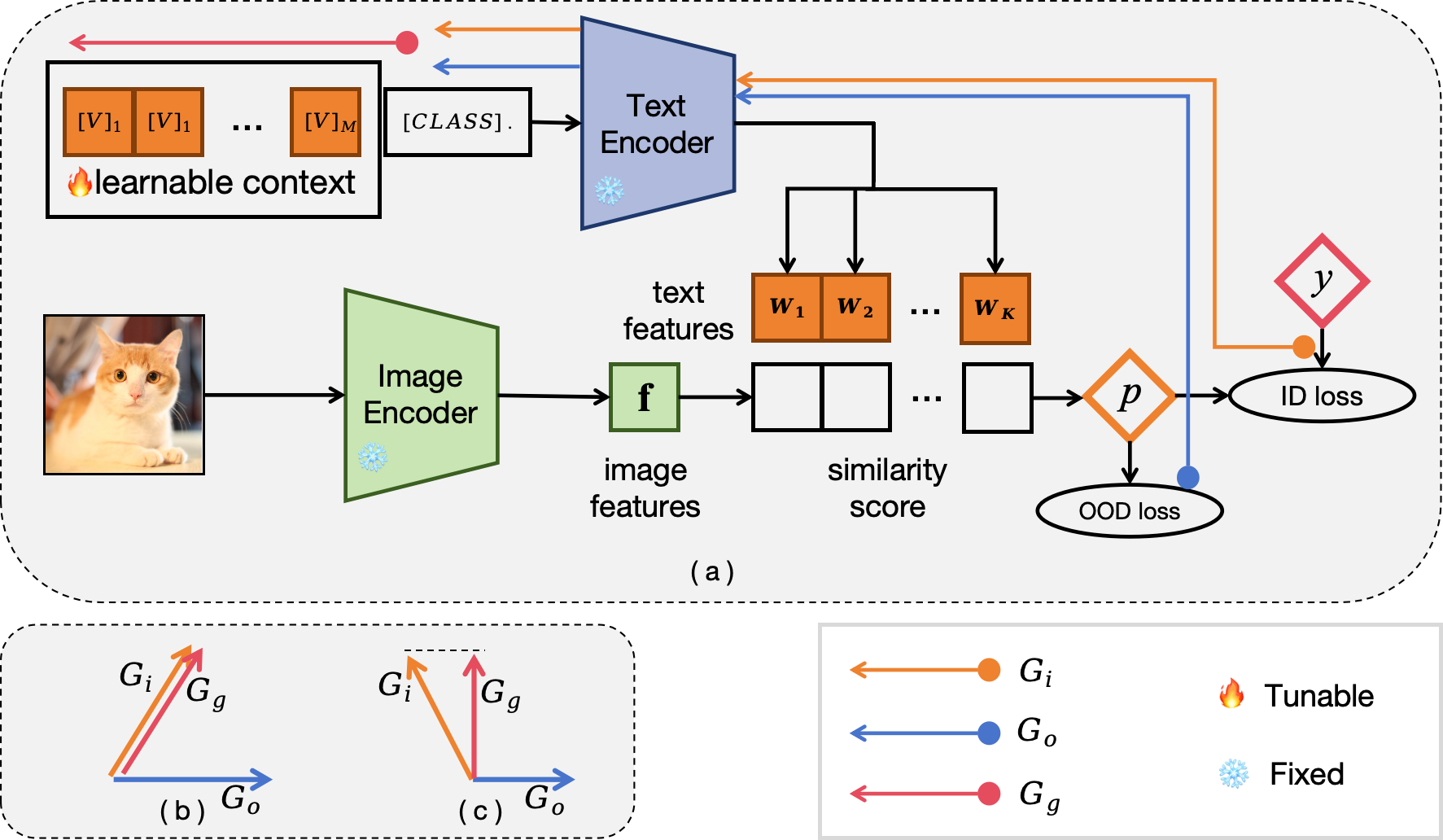}
        \caption{(a) Overall pipeline of the proposed Gradient Aligned Context Optimization(GaCoOp). (b) If $G_i$ is aligned with $G_o$, we set $G_{GaCoOp}$ as $G_i$. (c) If $G_i$ conflicts with $G_o$ (i.e., their angle is larger than $90^{\circ}$), we set $G_{gacoop}$ as the projection of $G_i$ on the orthogonal direction of $G_o$.}
        \label{fig-pipeline}
\end{figure}

\begin{table*}[bp]
  \caption{\textbf{Results on ImageNet OOD benchmarks.} We use CLIP-B/16 as a backbone, and use ImageNet-1K as ID. Bold values indicate the highest performance.}
  \label{exp_ood}
  \centering
  \footnotesize
  \setlength{\tabcolsep}{2.35mm}

\begin{tabular*}{\linewidth}{lcccccccccc}
\hline
\multirow[b]{2}{*}{\textbf{Method}} & \multicolumn{2}{c}{iNaturalist} & \multicolumn{2}{c}{SUN} & \multicolumn{2}{c}{Places} & \multicolumn{2}{c}{Texture} & \multicolumn{2}{c}{\textbf{Average}} \\
\cmidrule(l{1em}r{1.5em}){2-3}
\cmidrule(l{1em}r{1.5em}){4-5}
\cmidrule(l{1em}r{1.5em}){6-7}
\cmidrule(l{1em}r{1.5em}){8-9}
\cmidrule(l{1em}r{1.5em}){10-11}

& FPR95 $\downarrow$ & AUROC $\uparrow$ & FPR95 $\downarrow$ & AUROC $\uparrow$ & FPR95 $\downarrow$ & AUROC $\uparrow$ & FPR95 $\downarrow$ & AUROC $\uparrow$ & FPR95 $\downarrow$ & AUROC $\uparrow$ \\
\hline

\multicolumn{11}{c}{\textit{Full/Sub Data Fine-tune}} \\

ODIN~\cite{liang2018ODIN} & 30.22 & 94.65 & 54.04 & 87.17 & 55.06 & 85.54 & 51.67 & 87.85 & 47.75 & 88.80 \\

ViM~\cite{wang2022vim} & 32.19 & 93.16 & 54.01 & 87.19 & 60.67 & 83.75 & 53.94 & 87.18 & 50.20 & 87.82 \\

KNN~\cite{sun2022KNN} & 29.17 & 94.52 & 35.62 & 92.67 & 39.61 & 91.02 & 64.35 & 85.67 & 42.19 & 90.97 \\

NPOS~\cite{taonon-NPOS} & 16.58 & 96.19 & 43.77 & 90.44 & 45.27 & 89.44 & 46.12 & 88.80 & 37.93 & 91.22 \\

\hline

\multicolumn{11}{c}{\textit{Zero-shot}} \\

MCM~\cite{ming2022delving} & 30.94 & 94.61 & 37.67 & 92.56 & 44.76 & 89.76 & 57.91 & 86.10 & 42.82 & 90.76 \\

\hline
\multicolumn{11}{c}{\textit{One-shot}} \\

CoOp~\cite{zhou2022CoOp} & 43.38 & 91.26 & 38.53 & 91.95 & 46.68 & 89.09 & 50.64 & 87.83 & 44.81 & 90.03 \\

LoCoOp~\cite{miyai2024locoop} & 38.49 & 92.49 & 33.27 & 93.67 & 39.23 & 91.07 & 49.25 & 89.13 & 40.17 & 91.53 \\

ID-like~\cite{bai2024id-like} & 14.57 & 97.35 & 44.02 & 91.08 & 41.74 & 91.15 & 26.77 & \textbf{94.38}  & 31.78 & \textbf{93.47} \\
\rowcolor{gray!30}
GaCoOp (ours) & 15.47 & 96.69 & \textbf{24.58} & \textbf{94.57} & 35.17 & 91.05 & 43.12 & 89.29 & \textbf{29.59} & 92.90 \\

\multicolumn{11}{c}{\textit{Four-shot}}  \\

CoOp~\cite{zhou2022CoOp} & 35.36 & 92.60 & 37.06 & 92.27 & 45.38 & 89.15 & 43.74 & 89.68 & 40.39 & 90.92 \\

LoCoOp~\cite{miyai2024locoop}  & 29.45 & 93.93 & 33.06 & 93.24 & 41.13 & 90.32 & 44.15 & 90.54 & 36.95 & 92.01 \\

ID-like~\cite{bai2024id-like} & \textbf{8.98} & \textbf{98.19} & 42.03 & 91.64 & 44.00 & 90.57 & \textbf{25.27} & 94.32 & 30.07 & \textbf{93.68} \\

\rowcolor{gray!30}
GaCoOp (ours) & 15.42 & 96.69 & 25.19 & 94.43 & \textbf{34.43} & \textbf{91.40} & 42.57 & 89.95 & \textbf{29.40} & 93.12 \\
\hline
\end{tabular*}

\end{table*}

\section{Method}
\label{sec:pagestyle}

\subsection{Preliminaries}
\textbf{Contrastive language-image pre-training (CLIP)~\cite{radford2021learning}} has demonstrated remarkable zero-shot learning capabilities by leveraging contrastive learning with large-scale noisy image-text pairs. Via calculating the similarity between text and images, CLIP can perform effective classification ability~\cite{tang2024amu}. Given an input image $x_{in}$, the probability it belongs to class $n$ is $ p\left(y=n \mid \boldsymbol{x}^{\mathrm{in}}\right)$. The formulation is as follows:
\begin{equation}
    p\left(y=n \mid \boldsymbol{x}^{\mathrm{in}}\right)=\frac{\exp \left(\operatorname{sim}\left(\boldsymbol{f}{ }^{\text {in }}, \boldsymbol{g}_{n}\right) / \tau\right)}{\sum_{n=1}^{N} \exp \left(\operatorname{sim}\left(\boldsymbol{f}{ }^{\text {in }}, \boldsymbol{g}_{n}\right) / \tau\right)},
\end{equation}
where $\boldsymbol{f}^{in}$ and $\boldsymbol{g}_{n}$ are the corresponding image feature and text feature. The $\tau$ is the temperature parameter, and $N$ is the total number of classes.

\textbf{Local regularized context optimization (LoCoOp)~\cite{miyai2024locoop}} firstly adapt CLIP to out-of-distribution (OOD) detection task. 
Specifically, LoCoOp firstly computes the similarity between each region of the image and the text features and obtains the classification probability of each region relative to each class~\cite{zhou2022extract}. 
Then, it makes the entropy of  $p_{j}(y|x_{in})$ larger and enables the OOD region image features to be dissimilar to any ID text embedding with entropy maximization.
\begin{equation}
    \mathcal{L}_{\mathrm{ood}}=-H(p_j),
\end{equation}
where $H(\cdot)$ is the entropy function and $p_{j}$ denotes the classification prediction probability of the j-th ID-irrelevant region (e.g., backgrounds).
The final objective can be viewed as a combination of in-distribution (ID) classification optimization and out-of-distribution (OOD) regularization optimization~\cite{saito2020universal}. 
\begin{equation}
    \mathcal{L}=\mathcal{L}_{{\mathrm{coop}}}+\lambda\mathcal{L}_{{\mathrm{ood}}},
    \label{locoop_loss}
\end{equation}
where $\mathcal{L}_{{\mathrm{coop}}}$ is the standard classification loss based on the cross-entropy in ~\cite{zhou2022CoOp} and $\lambda$ is a hyperparameter.

When there is a bias in CLIP's recognition of ID-irrelevant regions, the $\mathcal{L}_{\mathrm{ood}}$ optimization to make the entropy of K ID-irrelevant region indices $p_{j}(y|x_{in})$ larger will lead to conflict with $\mathcal{L}_{{\mathrm{coop}}}$ optimization. This will limit the model's performance.

\begin{table}
\caption{ID  classification accuracy on ImageNet-1k}
\label{exp_id}
\footnotesize
\setlength{\tabcolsep}{1.8mm}
\begin{tabular}{lcccc}

\hline \text { Method } & \text { ID acc }  & \text { Zero-shot } & \text { Four-shot } & \text { Training time }\\
\hline 
\text { MCM~\cite{ming2022delving} } & 67.01  & \checkmark & & 0min\\
\text { CoOp~\cite{zhou2022CoOp} } & 69.38  & & \checkmark & 1h13min\\
\text { LoCoOp~\cite{miyai2024locoop} } & 69.19  & & \checkmark & 1h21min\\
\text { ID-like~\cite{bai2024id-like} } & 68.99  & & \checkmark &  23h51min\\
\rowcolor{gray!30}
\text { GaCoOp(ours) } & \textbf{69.63}  & & \checkmark & 1h48min\\

\hline
\end{tabular}
\end{table}

\subsection{Our Method}
To mitigate the conflict between ID classification optimization and OOD regularization optimization, we present a novel few-shot OOD detection prompt tuning method called Gradient aligned Context Optimization (GaCoOp). 

Specifically, we define that the ID classification optimization direction $G_{i}$ is the gradient of the cross-entropy classification loss ($L_{coop}$).
Similarly, we compute the OOD regularization optimization direction $G_{o}$ using the gradient of OOD regularization loss ($L_{ood}$). We name $G_{o}$ as the OOD regularization direction.
Inspired by~\cite{zhu2023promptgrad}, we decompose the ID classification direction $G_{i}$ into 1) orthogonal direction, where the projection of $G_{i}$ orthogonal to the $G_{o}$. This denotes the non-conflicting knowledge; 
and 2) parallel direction, where the projection of $G_{i}$ parallel to the $G_{o}$. This denotes the potentially conflicting knowledge of OOD regularization.
Then, the orthogonal direction satisfies that it can not override the $G_{i}$ as any two orthogonal vectors can be transformed into two non-conflicting base vectors.
The parallel direction satisfies one of the two directions: 1) the same as the OOD regularization direction, which indicates that the update is aligned to the ID classification knowledge, and 2) the opposite of OOD regularization direction, indicating a conflicting update that should be discarded.

Without loss of generality, these properties lead the two cases for $G_{i}$ and $G_{o}$
(1) As shown in Fig.~\ref{fig-pipeline}(b), their optimization angle is smaller than $90^{\circ}$ (acute angle), which indicates that the optimization direction of ID classification does not conflict with OOD regularization. In this case, we safely set the updated gradient direction $G_{gacoop}$ as $G_{i}$. (2) As shown in Fig.~\ref{fig-pipeline}(c), their optimization angle is larger than $90^{\circ}$ (obtuse angle). We project the $G_{i}$ to the orthogonal direction of $G_{o}$ to optimize the model.
The optimization strategy of our GaCoOp can be formulated as follows:
\begin{equation}
    \boldsymbol{G}_{\text {gacoop }}=\left\{\begin{array}{ll}
\boldsymbol{G}_{\mathrm{i}}, & \text { if } \boldsymbol{G}_{\mathrm{i}} \cdot \boldsymbol{G}_{\mathrm{o}} \geq 0 \\
\boldsymbol{G}_{\mathrm{i}}- \frac{\boldsymbol{G}_{\mathrm{i}} \cdot \boldsymbol{G}_{\mathrm{o}}}{\left\|\boldsymbol{G}_{\mathrm{o}}\right\|^{2}} \boldsymbol{G}_{\mathrm{o}}, & \text { otherwise. }
\end{array}\right.
\end{equation}
Fig.~\ref{fig-pipeline}(a) shows the overall pipeline of our method. Instead of updating the context vectors using the loss function in (\ref{locoop_loss}), we regard the optimization of the OOD regularization optimization direction as a reference, employing gradient projection to mitigate the issue where the ID and OOD optimization directions conflict due to biases in VLMs. 
We believe a perfect unbiased classification model is also a good OOD detection model.

\section{Experiment}
\label{sec:typestyle}

\subsection{Experimental Detail}

\textbf{Datasets.} Same as prior works~\cite{ming2022delving,miyai2024locoop,bai2024id-like}, we evaluate our method on the large-scale ImageNet OOD detection benchmarks. Specifically, we use the ImageNet-1K~\cite{deng2009imagenet} dataset as the ID data~\cite{bai2024id-like}, and for OOD datasets, we adopt four datasets, including subsets of iNaturalist~\cite{van2018inaturalist}, SUN~\cite{xiao2010sun}, Places~\cite{zhou2017places}, and TEXTURE~\cite{cimpoi2014describing}. During the few-shot training, We adopt the same settings as the recent method ID-like~\cite{bai2024id-like}. During the test phase, we evaluate our method on the whole test set. 

\textbf{Baselines. }In the experiments, to assess the effectiveness of our method in mitigating conflict, we conducted experiments on both in-distribution (ID) classification and out-of-distribution (OOD) detection. For ID classification accuracy, we compare with MCM~\cite{ming2022delving}, CoOp~\cite{zhou2022CoOp}, LoCoOp~\cite{miyai2024locoop}, ID-like~\cite{bai2024id-like}, and for OOD detection, similar with~\cite{bai2024id-like}, we compare with zero-shot detection methods, fully-supervised detection methods, and the baseline prompt learning method. For zero-shot out-of-distribution (OOD) detection methods, we employ MCM~\cite{ming2022delving}. For fully-supervised detection methods, we compare with ODIN~\cite{liang2018ODIN}, ViM~\cite{wang2022vim}, KNN~\cite{sun2022KNN} and NPOS~\cite{taonon-NPOS}. For prompt learning methods, we use LoCoOp~\cite{miyai2024locoop}, which is the representative work and the baseline for our GaCoOp. We also compared our method with the recent powerful approach: ID-like~\cite{bai2024id-like}. For fairness, all methods are trained using the same pre-trained model (CLIP/ViT-B/16), and we reproduce some results from LoCoOp~\cite{miyai2024locoop} and ID-like~\cite{bai2024id-like}.

\textbf{Metrics. }For evaluation, we utilize the following metrics: (1) the false positive rate (FPR95) of out-of-distribution (OOD) samples when the true positive rate of in-distribution (ID) samples is at 95$\%$, (2) the area under the receiver operating characteristic curve (AUROC), and (3) in-distribution (ID) classification accuracy (ID ACC).

\textbf{Implementation Details. }Same as prior works, we use CLIP~\cite{radford2021learning} ViT-B/16 from the official repository as backbone. 
And for hyperparameters, such as training epochs (50), learning rate (0.002), batch size (32), and token lengths (N=16), remain consistent with those of baseline~\cite{miyai2024locoop}. In all experiments, we evaluate on NVIDIA RTX3090 GPUs.

\begin{figure}[bp]
\begin{minipage}[b]{.48\linewidth}
  \centering
  \centerline{\includegraphics[width=4.0cm]{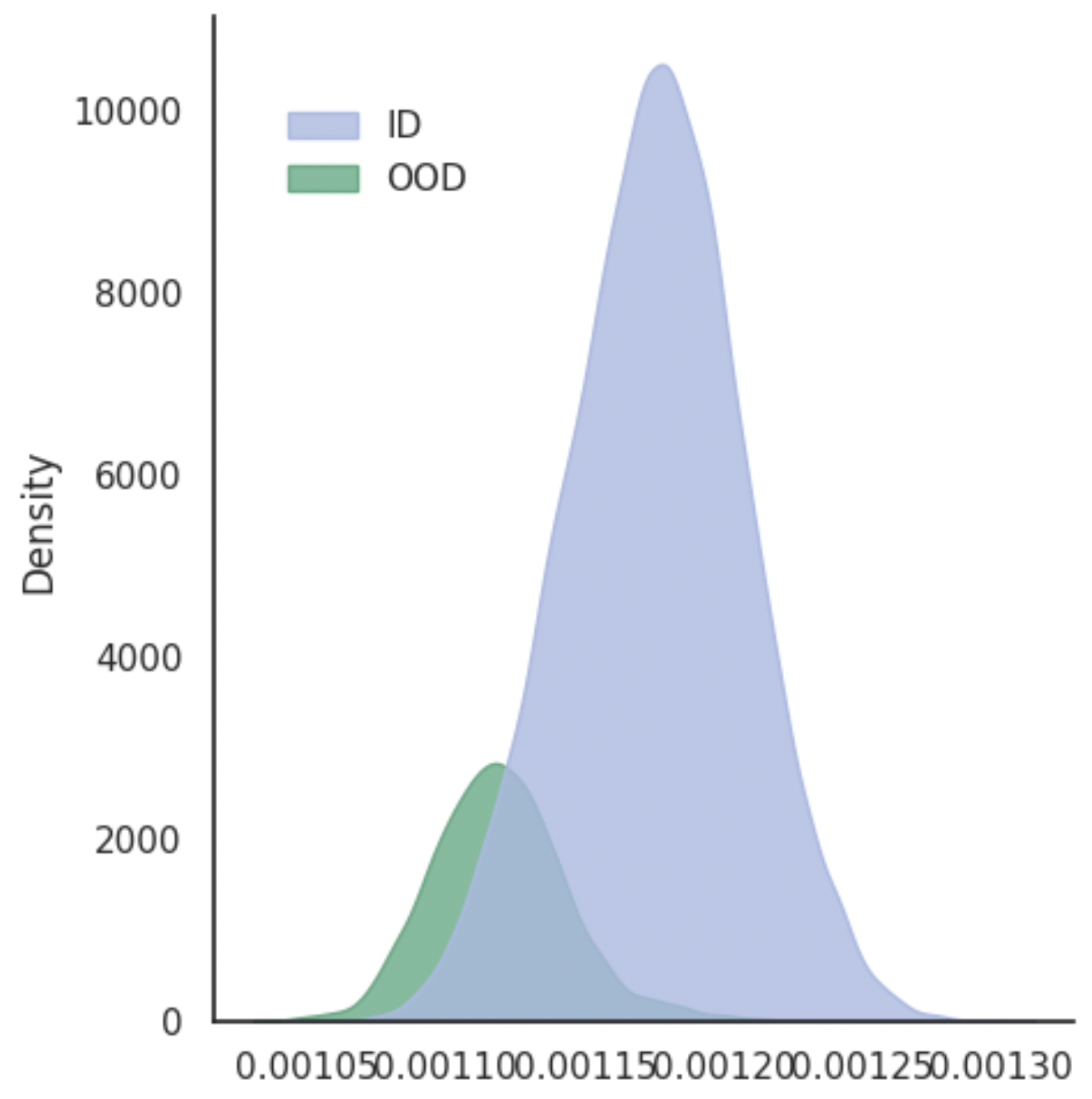}}
  \centerline{(a) CoOp}\medskip
\end{minipage}
\hfill
\begin{minipage}[b]{0.48\linewidth}
  \centering
  \centerline{\includegraphics[width=4.0cm]{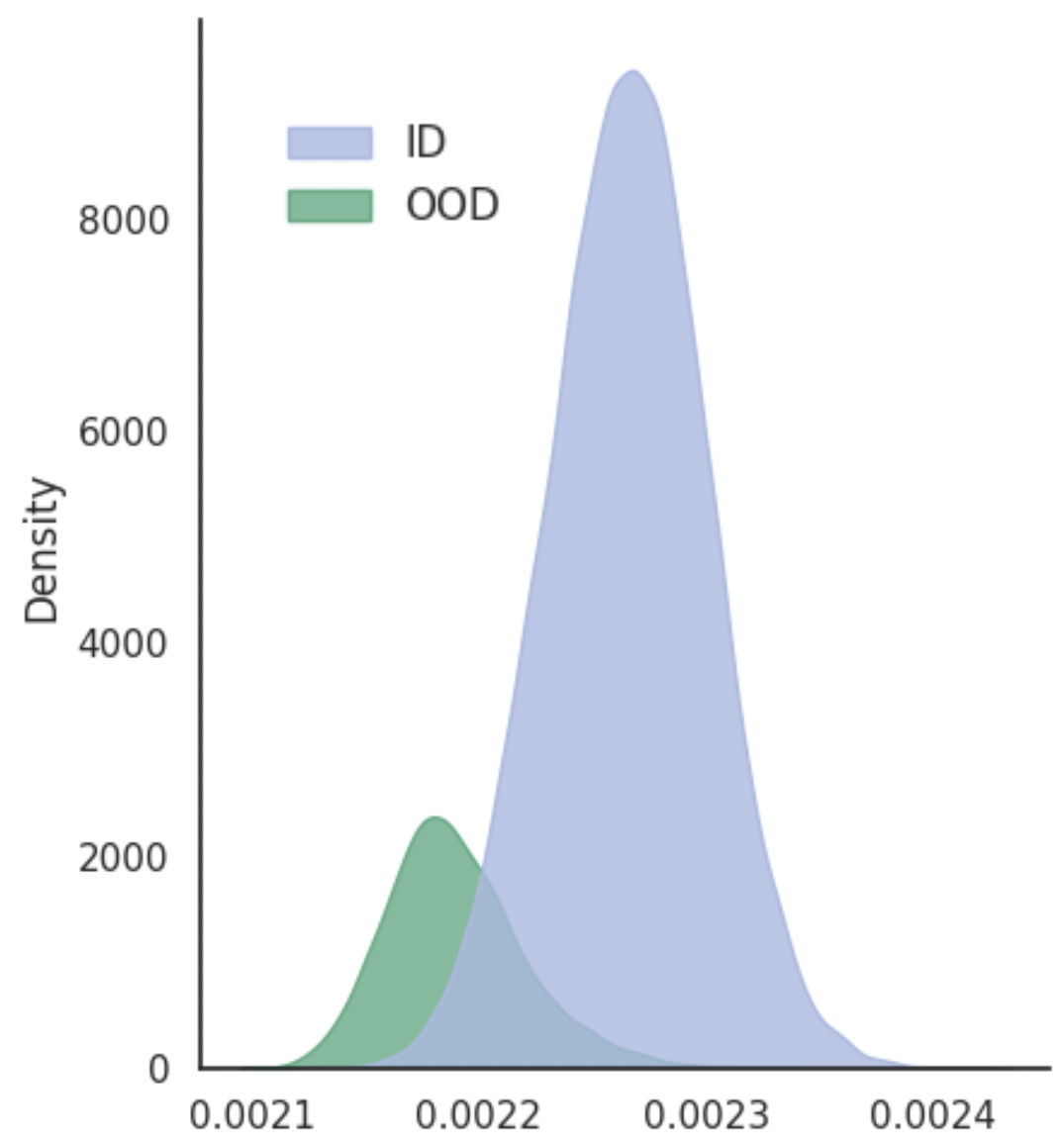}}
  \centerline{(b) Ours}\medskip
\end{minipage}
\caption{Density of the obtained ID and OOD score on SUN dataset.}
\label{fig:ablation}
\end{figure}

\subsection{Main results}

\textbf{Results under ID classification. } As shown in the Table~\ref{exp_id}, our method achieves better results with less training overhead. Specifically, compared to similar OOD detection methods, our method not only achieves the best classification performance but also maintains an acceptable training overhead, taking only about 30min longer than the optimal time. When compared with the method designed for classification, our approach also surpassed CoOp by 0.25\% on ImageNed-1k.

\textbf{Results under OOD detection. } As shown in the Table~\ref{exp_ood}, our method also achieves better out-of-distribution (OOD) detection performance, outperforming most comparisons. Specifically, when compared with the zero-shot method, on average, our approach surpasses the competing method MCM by more than 2.14\% in AUC and around 13.23\% in FPR95 in the one-shot scenario. When compared with the prompt learning-based methods, our approach surpasses the baseline method LoCoOp by more than 1.11\% in AUC and around 7.55\% in FPR95 in the four-shot scenario. Although our method does not achieve the best results in AUC, it is worth noting that our method learns only a single prompt, yet it still surpasses recent method ID-like that learns two prompts, reducing the FPR95 by 2.19\% (one-shot) and 0.67\% (four-shot). 

Moreover, Combining Table~\ref{exp_id} and Table~\ref{exp_ood}, we can observe that although the recent method ID-like achieves good results in OOD detection, its performance in ID classification is not very good. 
We explain this from the perspective of gradient optimization, suggesting that the bias inherent in CLIP itself leads to a conflict between optimizing for ID classification and OOD detection. Our method can help mitigate this conflict, resulting in better ID classification and OOD detection performance.

\subsection{Ablation study}

\textbf{Effectiveness with CNN architectures.} We also examine the effectiveness of GaCoOp with CNN architectures. We use a Vision Transformer architecture as the backbone following existing work~\cite{ming2022delving,miyai2024locoop,bai2024id-like} in the main experiments. However, CNN-based methods are also a popular research topic in the field of OOD detection. Following this practice, we use ResNet-50 as the CNN backbone and evaluate the effectiveness of our method. Table~\ref{exp_res50} shows the four-shot results with CLIP-ResNet-50 and MCM~\cite{ming2022delving} is a zero-shot method among them. The ID accuracy result is derived from the ImageNet-1K dataset and the reported FPR95 and AUROC are the average across four datasets. We find that GaCoOp can also achieve better results.

\begin{table}
\caption{Comparison results with CLIP-ResNet-50.}
\label{exp_res50}
\footnotesize
\setlength{\tabcolsep}{1.8mm}
\begin{tabular}{lcccc}

\hline 
\text { Method } & \text { ID acc }   & \text { FPR95 $\downarrow$ } & \text { AUROC $\uparrow$ } & \text { Training time }\\
\hline 
\text { MCM~\cite{ming2022delving} } & 58.18  & 49.66  &  89.00   & 0min\\
\text { CoOp~\cite{zhou2022CoOp} } & 60.01  & 48.15 & 89.22 & 1h13min\\
\text { LoCoOp~\cite{miyai2024locoop} } & 60.49  & 45.62 & 90.74   & 1h14min\\
\text { ID-like~\cite{bai2024id-like} } & 60.34  & 39.83 & 91.29  &  17h57min\\
\rowcolor{gray!30}
\text { GaCoOp(ours) } & \textbf{60.50}  & \textbf{33.74} & \textbf{91.62} & 1h41min\\

\hline
\end{tabular}
\end{table}

\section{Conclusion}
\label{sec:conclusion}
We present GaCoOp, a novel approach for few-shot OOD detection. It alleviates the potential optimization conflict between the ID optimization direction and the OOD regularization direction with gradient projection. Our empirical results reveal that our GaCoOp can achieve both better classification performance and OOD detection capability after mitigating conflicts. 
This also further motivates us to consider the biases hidden in pre-trained models.

\vfill\pagebreak

\bibliographystyle{IEEE_ref_VLMsTTA/IEEE_ref_bib}
\bibliography{IEEE_ref_VLMsTTA/IEEE_ref}

\begin{thebibliography}{10}
\providecommand{\url}[1]{#1}
\csname url@samestyle\endcsname
\providecommand{\newblock}{\relax}
\providecommand{\bibinfo}[2]{#2}
\providecommand{\BIBentrySTDinterwordspacing}{\spaceskip=0pt\relax}
\providecommand{\BIBentryALTinterwordstretchfactor}{4}
\providecommand{\BIBentryALTinterwordspacing}{\spaceskip=\fontdimen2\font plus
\BIBentryALTinterwordstretchfactor\fontdimen3\font minus \fontdimen4\font\relax}
\providecommand{\BIBforeignlanguage}[2]{{%
\expandafter\ifx\csname l@#1\endcsname\relax
\typeout{** WARNING: IEEEtran.bst: No hyphenation pattern has been}%
\typeout{** loaded for the language `#1'. Using the pattern for}%
\typeout{** the default language instead.}%
\else
\language=\csname l@#1\endcsname
\fi
#2}}
\providecommand{\BIBdecl}{\relax}
\BIBdecl

\bibitem{shu2023clipood}
Y.~Shu, X.~Guo, J.~Wu, X.~Wang, J.~Wang, and M.~Long, ``Clipood: Generalizing clip to out-of-distributions,'' in \emph{International Conference on Machine Learning}.\hskip 1em plus 0.5em minus 0.4em\relax PMLR, 2023, pp. 31\,716--31\,731.

\bibitem{hendrycks2016baseline}
D.~Hendrycks and K.~Gimpel, ``A baseline for detecting misclassified and out-of-distribution examples in neural networks,'' \emph{arXiv preprint arXiv:1610.02136}, 2016.

\bibitem{radford2021learning}
A.~Radford, J.~W. Kim, C.~Hallacy, A.~Ramesh, G.~Goh, S.~Agarwal, G.~Sastry, A.~Askell, P.~Mishkin, J.~Clark \emph{et~al.}, ``Learning transferable visual models from natural language supervision,'' in \emph{International conference on machine learning}.\hskip 1em plus 0.5em minus 0.4em\relax PMLR, 2021, pp. 8748--8763.

\bibitem{ming2022delving}
Y.~Ming, Z.~Cai, J.~Gu, Y.~Sun, W.~Li, and Y.~Li, ``Delving into out-of-distribution detection with vision-language representations,'' \emph{Advances in neural information processing systems}, vol.~35, pp. 35\,087--35\,102, 2022.

\bibitem{liang2018ODIN}
S.~Liang, Y.~Li, and R.~Srikant, ``Enhancing the reliability of out-of-distribution image detection in neural networks,'' in \emph{6th International Conference on Learning Representations, ICLR 2018}, 2018.

\bibitem{wang2022vim}
H.~Wang, Z.~Li, L.~Feng, and W.~Zhang, ``Vim: Out-of-distribution with virtual-logit matching,'' in \emph{Proceedings of the IEEE/CVF conference on computer vision and pattern recognition}, 2022, pp. 4921--4930.

\bibitem{sun2022KNN}
Y.~Sun, Y.~Ming, X.~Zhu, and Y.~Li, ``Out-of-distribution detection with deep nearest neighbors,'' in \emph{International Conference on Machine Learning}.\hskip 1em plus 0.5em minus 0.4em\relax PMLR, 2022, pp. 20\,827--20\,840.

\bibitem{taonon-NPOS}
L.~Tao, X.~Du, J.~Zhu, and Y.~Li, ``Non-parametric outlier synthesis,'' in \emph{The Eleventh International Conference on Learning Representations}, 2023.

\bibitem{miyai2024locoop}
A.~Miyai, Q.~Yu, G.~Irie, and K.~Aizawa, ``Locoop: Few-shot out-of-distribution detection via prompt learning,'' \emph{Advances in Neural Information Processing Systems}, vol.~36, 2024.

\bibitem{bai2024id-like}
Y.~Bai, Z.~Han, B.~Cao, X.~Jiang, Q.~Hu, and C.~Zhang, ``Id-like prompt learning for few-shot out-of-distribution detection,'' in \emph{Proceedings of the IEEE/CVF Conference on Computer Vision and Pattern Recognition}, 2024, pp. 17\,480--17\,489.

\bibitem{yang2023invariant}
M.~Yang, Z.~Fang, Y.~Zhang, Y.~Du, F.~Liu, J.-F. Ton, J.~Wang, and J.~Wang, ``Invariant learning via probability of sufficient and necessary causes,'' \emph{Advances in Neural Information Processing Systems}, vol.~36, pp. 79\,832--79\,857, 2023.

\bibitem{song2024fd-align}
K.~Song, H.~Ma, B.~Zou, H.~Zhang, and W.~Huang, ``Fd-align: feature discrimination alignment for fine-tuning pre-trained models in few-shot learning,'' \emph{Advances in Neural Information Processing Systems}, vol.~36, 2024.

\bibitem{wang2024learnclip-bias}
J.~Wang and G.~Kang, ``Learn to rectify the bias of clip for unsupervised semantic segmentation,'' in \emph{Proceedings of the IEEE/CVF Conference on Computer Vision and Pattern Recognition}, 2024, pp. 4102--4112.

\bibitem{li2019visualbert}
L.~H. Li, M.~Yatskar, D.~Yin, C.-J. Hsieh, and K.-W. Chang, ``Visualbert: A simple and performant baseline for vision and language,'' \emph{arXiv preprint arXiv:1908.03557}, 2019.

\bibitem{kim2021vilt}
W.~Kim, B.~Son, and I.~Kim, ``Vilt: Vision-and-language transformer without convolution or region supervision,'' in \emph{International conference on machine learning}.\hskip 1em plus 0.5em minus 0.4em\relax PMLR, 2021, pp. 5583--5594.

\bibitem{yaofilip}
L.~Yao, R.~Huang, L.~Hou, G.~Lu, M.~Niu, H.~Xu, X.~Liang, Z.~Li, X.~Jiang, and C.~Xu, ``Filip: Fine-grained interactive language-image pre-training,'' in \emph{International Conference on Learning Representations}, 2021.

\bibitem{li2022blip}
J.~Li, D.~Li, C.~Xiong, and S.~Hoi, ``Blip: Bootstrapping language-image pre-training for unified vision-language understanding and generation,'' in \emph{International conference on machine learning}.\hskip 1em plus 0.5em minus 0.4em\relax PMLR, 2022, pp. 12\,888--12\,900.

\bibitem{zhou2022CoOp}
K.~Zhou, J.~Yang, C.~C. Loy, and Z.~Liu, ``Learning to prompt for vision-language models,'' \emph{International Journal of Computer Vision}, vol. 130, no.~9, pp. 2337--2348, 2022.

\bibitem{jia2022visualprompt}
M.~Jia, L.~Tang, B.-C. Chen, C.~Cardie, S.~Belongie, B.~Hariharan, and S.-N. Lim, ``Visual prompt tuning,'' in \emph{European Conference on Computer Vision}.\hskip 1em plus 0.5em minus 0.4em\relax Springer, 2022, pp. 709--727.

\bibitem{khattak2023maple}
M.~U. Khattak, H.~Rasheed, M.~Maaz, S.~Khan, and F.~S. Khan, ``Maple: Multi-modal prompt learning,'' in \emph{Proceedings of the IEEE/CVF Conference on Computer Vision and Pattern Recognition}, 2023, pp. 19\,113--19\,122.

\bibitem{pratt2023does}
S.~Pratt, I.~Covert, R.~Liu, and A.~Farhadi, ``What does a platypus look like? generating customized prompts for zero-shot image classification,'' in \emph{Proceedings of the IEEE/CVF International Conference on Computer Vision}, 2023, pp. 15\,691--15\,701.

\bibitem{fort2021exploring}
S.~Fort, J.~Ren, and B.~Lakshminarayanan, ``Exploring the limits of out-of-distribution detection,'' \emph{Advances in Neural Information Processing Systems}, vol.~34, pp. 7068--7081, 2021.

\bibitem{du-vos}
X.~Du, Z.~Wang, M.~Cai, and Y.~Li, ``Vos: Learning what you don't know by virtual outlier synthesis,'' in \emph{International Conference on Learning Representations}, 2022.

\bibitem{tang2024amu}
Y.~Tang, Z.~Lin, Q.~Wang, P.~Zhu, and Q.~Hu, ``Amu-tuning: Effective logit bias for clip-based few-shot learning,'' in \emph{Proceedings of the IEEE/CVF Conference on Computer Vision and Pattern Recognition}, 2024, pp. 23\,323--23\,333.

\bibitem{zhou2022extract}
C.~Zhou, C.~C. Loy, and B.~Dai, ``Extract free dense labels from clip,'' in \emph{European Conference on Computer Vision}.\hskip 1em plus 0.5em minus 0.4em\relax Springer, 2022, pp. 696--712.

\bibitem{saito2020universal}
K.~Saito, D.~Kim, S.~Sclaroff, and K.~Saenko, ``Universal domain adaptation through self supervision,'' \emph{Advances in neural information processing systems}, vol.~33, pp. 16\,282--16\,292, 2020.

\bibitem{zhu2023promptgrad}
B.~Zhu, Y.~Niu, Y.~Han, Y.~Wu, and H.~Zhang, ``Prompt-aligned gradient for prompt tuning,'' in \emph{Proceedings of the IEEE/CVF International Conference on Computer Vision}, 2023, pp. 15\,659--15\,669.

\bibitem{deng2009imagenet}
J.~Deng, W.~Dong, R.~Socher, L.-J. Li, K.~Li, and L.~Fei-Fei, ``Imagenet: A large-scale hierarchical image database,'' in \emph{2009 IEEE conference on computer vision and pattern recognition}.\hskip 1em plus 0.5em minus 0.4em\relax Ieee, 2009, pp. 248--255.

\bibitem{van2018inaturalist}
G.~Van~Horn, O.~Mac~Aodha, Y.~Song, Y.~Cui, C.~Sun, A.~Shepard, H.~Adam, P.~Perona, and S.~Belongie, ``The inaturalist species classification and detection dataset,'' in \emph{Proceedings of the IEEE conference on computer vision and pattern recognition}, 2018, pp. 8769--8778.

\bibitem{xiao2010sun}
J.~Xiao, J.~Hays, K.~A. Ehinger, A.~Oliva, and A.~Torralba, ``Sun database: Large-scale scene recognition from abbey to zoo,'' in \emph{2010 IEEE computer society conference on computer vision and pattern recognition}.\hskip 1em plus 0.5em minus 0.4em\relax IEEE, 2010, pp. 3485--3492.

\bibitem{zhou2017places}
B.~Zhou, A.~Lapedriza, A.~Khosla, A.~Oliva, and A.~Torralba, ``Places: A 10 million image database for scene recognition,'' \emph{IEEE transactions on pattern analysis and machine intelligence}, vol.~40, no.~6, pp. 1452--1464, 2017.

\bibitem{cimpoi2014describing}
M.~Cimpoi, S.~Maji, I.~Kokkinos, S.~Mohamed, and A.~Vedaldi, ``Describing textures in the wild,'' in \emph{Proceedings of the IEEE conference on computer vision and pattern recognition}, 2014, pp. 3606--3613.

\end{thebibliography}

\end{document}